# An Analysis of the Differences Among Regional Varieties of Chinese in Malay Archipelago

Nankai Lin[1], Sihui Fu[2], Hongyan Wu[2], Shengyi Jiang[2](✉)

**Abstract**
Chinese features prominently in the Chinese communities located in the nations of Malay Archipelago. In these countries, Chinese has undergone the process of adjustment to the local languages and cultures, which leads to the occurrence of a Chinese variant in each country. In this paper, we conducted a quantitative analysis on Chinese news texts collected from five Malay Archipelago nations, namely Indonesia, Malaysia, Singapore, Philippines and Brunei, trying to figure out their differences with the texts written in modern standard Chinese from a lexical and syntactic perspective. The statistical results show that the Chinese variants used in these five nations are quite different, diverging from their modern Chinese mainland counterpart. Meanwhile, we managed to extract and classify several featured Chinese words used in each nation. All these discrepancies reflect how Chinese evolves overseas, and demonstrate the profound impact rom local societies and cultures on the development of Chinese.

## Introduction

In a broad sense, Chinese refers to a common Chinese language based on Mandarin Chinese (Ai, 2011). With immigration and external communication, Chinese used in many countries features prominently in the Chinese communities especially located in the nations of Southeast Asia. Chinese has undergone the process of adjustment to the local policies and cultures in different countries, which leads to the occurrence of a Chinese variant in each country. For instance, Brunei's Chinese education has struggled under external constraints of education policy with priority in English and Malaysia adopted by the western colonial government and the Malaysian government. Moreover, many problems still exist in Brunei's Chinese schools——the leading platform of Chinese education. Restrictions on textbooks and faculty have also had an undesirable impact on the development of the Chinese in Brunei (Wen, 2015). A prominent feature of Brunei's language policy is the emphasis on standard Malay, a focus on English and support for Malay in Brunei (Zhang and Guo, 2016). Thus, Chinese is underappreciated, which hinders the development of Chinese in Brunei. In the Philippines, Chinese education has undergone the process of germination, prosperity, stagnation as well as recovery after the war. However, the implementation of the Filipino policy on Chinese education during the reign of Marcus seriously inhibited the development of Chinese education (Sui, 2016), resulting in the lack of Chinese education resources. Dominated by a single grammatical translation in teaching, Chinese education under the supervision of the church cannot develop freely. Following the traditional teaching mode (Fan, 2020), the level of Chinese in the Philippines gradually decreases. In Indonesia, Chinese are not proficient in Chinese due to the fact that since the 1965 political coup, Indonesian authorities have closed overseas Chinese schools and banned the public display of Chinese art and culture, including the employment of Chinese. Existing research indicates that Chinese language competence of Chinese in Indonesia is the weakest in comparison to that in Malaysia, Indonesia and Thailand (Zhou and An, 2019). Meanwhile, affected by various languages or dialects, Indonesian Chinese has some variations or errors compared with Mandarin Chinese (Zhang, 2010). Meanwhile, language education norms are also a significant factor causing the differentiation of Chinese. In the process of promoting and enhancing the Malaysian language as a national language by the Malaysian government, Malaysian Chinese groups have made unremitting efforts to inherit the native Chinese language. Now in Malaysia, there is a relatively integrated system from primary school, middle school to college (Ke, 2009). Chinese, as a common language of Chinese in Singapore, enjoys the same dominant status in Singapore as mandarin does in China. Use of the Chinese language in different communities is multiple and complex owing to differences in language skills, personal assessment tendency and family language planning (He and Wu, 2021). Moreover, due to the discrepancies in historical background, economic development, cultural education and demographic structure between Singapore and China, and the lack of communication between Singaporean Chinese and Mandarin, Singaporean Chinese based on Mandarin is different from Mandarin, especially in terms of vocabulary (Qi, 2014).

Therefore, there are significant differences in the development of Chinese in the countries of Malay Archipelago despite of the same origin with Mandarin in the Mainland. Quantitative and qualitative analysis of these discrepancies is conducive to exploring the causes and understanding the evolution of overseas Chinese. Nevertheless, few scholars have systematically figured out the differences with the texts written in modern standard Chinese, from the perspective of metrology. In this paper, employing the quantitative analysis, we attempted to explore Chinese news texts collected from five Malay Archipelago nations, namely Indonesia, Malaysia, Singapore, Philippines and Brunei to enrich the quantitative research on Chinese differences, contributing to a better understanding of Chinese in overseas Chinese newspapers and opening up new ideas for further research.

**Related work**

In the existing studies, some scholars have studied the differences in Chinese in different regions based on news texts, focusing mainly on the differences of Chinese used in Taiwan and Macao and Mandarin in the Mainland. Referring to Biber's multi-feature and multi-dimensional language analysis method, Xu and Li (2019) quantitatively analyze the expressions in the sports news between both sides of Taiwan Strait, from the perspectives of standard type token ratio, lexical density, the coverage rate of word frequency, high-frequency words, low-frequency words, the distribution of word classes, word length, sentence length and sentence category. Yang (2011) analyzes the types of different words collected from five newspapers between both sides of the Taiwan Strait and the reasons for the differences and finds the phenomenon of word fusion in media languages. Yao and Huang (2014) make a quantitative comparison of the usage of words between Macao and the Mainland based on news corpus, and study the driving factors of the variance by taking the auxiliary word "了(le)" as a case. On the other hand, many scholars have also researched the differences between the Chinese in the nations of the Malay Archipelago and that in the Mainland. Based on the quantitative statistics of Chinese sports news corpus both at home and abroad. Zhang et al. (2015) attempt to find the similarities and differences in the usage of words. Aiming to explore the differences between Singaporean Chinese and Mandarin Chinese, as well as the reasons for the differences, Chen (2008) conducts a systematic study on Singaporean Chinese words. Chanicha (2016) selects corpus different from Mandarin from a large number of Thai Chinese newspapers and Chinese network media and compares Thai Chinese with Mandarin concerned with vocabulary, grammar and style.

Taking five Southeast Asian countries (Thailand, Vietnam, Myanmar, Malaysia and Indonesia) as the research object, Wang (2018) aims to explore the correlation between Chinese communication and cognition of Chinese national image. Based on the corpus of Chinese in Singapore and Mandarin, Qi (2014) combines previous research results and divides the vocabulary into four categories for comparison, namely homographs with different meanings, stereographs with different meanings, homographs with different colors and usages, and their own unique words and catchwords. Simultaneously, Qi analyzes the lexical differences between Singaporean Chinese and Mandarin and investigates the reasons for the differences. Jia and Xu (2005) classify the differences between Singaporean Chinese-specific words and Mandarin words in terms of structure and origin into five categories: differences in the choice of word-forming elements, differences in abbreviation or non-abbreviation, differences in the taking of old words, differences in the absorption of Cantonese dialect words, and differences in the translation and absorption of foreign words. Sang (2020) analyzes the lexical differences between Indonesian Chinese and written Mandarin concerning morphological and semantic differences, category divergence, word collocation differences, pragmatic differences and Indonesian Chinese-specific words, and analyses the reasons for these differences from the two aspects of environment and history.

Generally speaking, there are two problems in the analysis of the differences between the Chinese in the nations of Malay Archipelago and that in the Mainland.

(1) Studies have largely been conducted on the Chinese language in individual countries only, with few systematic analysis of Chinese language differences across multiple countries.

(2) Related research has mainly focused on qualitative analysis, and few research has been conducted on Chinese language differences employing the quantitative analysis.

## Research Methodology

In this paper, we conducted a quantitative analysis on Chinese news texts collected from five Malay Archipelago nations, namely Indonesia, Malaysia, Singapore, Philippines and Brunei. Comparing different metrics, we attempt to investigate the richness and comprehension difficulties of Chinese language texts in these five countries to explore the differences in Chinese texts in different countries.

### Research Material

Table 1 The data size from each news source

| Country | Number of example words | Source |
| --- | --- | --- |
| Indonesia | 43482 | Indonesia Business (http://www.shangbaoindonesia.com/) |
|  |  | International Business Daily Indonesia (http://www.ibtimes.com.cn) |
| Malaysia | 48922 | Guanghua Daily (https://www.kwongwah.com.my/) |
|  |  | Shihua Daily (http://news.seehua.com/) |
|  |  | Sing Tao Daily (https://www.sinchew.com.my/) |
| Brunei | 49670 | Chinese Embassy in Brunei (http://bn.china-embassy.org/chn/) |
| Singapore | 49179 | United Morning News (http://www.zaobao.com/) |
| Philippines | 43840 | United Daily News (http://www.unitednews.net.ph) |
| China | 50424 | Agriculture Nanjing University Humanities and Social Computing Research Center "People's Daily" corpus |

We select influential Chinese news websites for each country and crawl all categories of news from 2004 to 2019 to study the usage of Chinese in five countries. Simultaneously, we adopt the Chinese word segmentation datasets[3] released by the Center for Humanities and Social Computing of Nanjing Agricultural University as the representative corpus of Chinese in the Mainland, aiming to facilitate our subsequent comparative study of the language differences between the nations of the Malay Archipelago and the Mainland regarding Chinese news reports. The dataset is derived from the news data of *People's Daily* in January 2018, and the original text realizes sentence extraction and manual word segmentation according to a pre-defined annotation specification (Huang and Wang, 2019). Therefore, we generally follow the principles of homogeneity and co-occurrence in corpus selection. Subsequently, considering the large variation in the length of the reported texts between different media, we randomly select 2000 sentences from each news source and then employ the Jieba[4] text segmentation tool for automatic word segmentation. To reduce the error rate, we also manually verified and corrected the word segmentation results, and finally obtain the relevant data of the corpus as shown in Table 1.

**Lexical Metrics**

The study of differences in quantitative indicators concerning words is the basis for the study of textual differences. In practical research, it is indispensable to explore the differences between languages by employing different metrics to uncover the metric characteristics of words in one language on different statistical parameters (Liu and Pan, 2015; Pan 2015). This paper investigates differences at the lexical level in terms of several metrics, where word type count, type token ratio and hapax legomena are used to describe the lexical richness of the corpus, $H$ value, $R$ value and $a$ value are used to describe the textual representation richness of the corpus, and the number of monosyllabic free morphemes describes the textual complexity.

    a)   **The number of word type and type token ratio**: The number of word type refers to the number of words that do not recur in the text. In order to eliminate the effect of text length on the number of word types, the concept of type token ratio has also been introduced. Type token ratio is the ratio of the number of word types in a text to the number of word examples, which reflects the richness of a text's vocabulary.

    b)   **The number of monosyllabic free morphemes**: A free morpheme consisting of a single phoneme is a monosyllabic free morpheme. In Chinese language, one Chinese character corresponds to one phoneme, hence, each word with only one character after word segmentation is a monosyllabic free morpheme. Liu (2018) indicates that the historical evolution of Chinese word length is a process of polysyllabification. Chen (2016) also suggests that the average word length increased during the evolution of Chinese. The closer the corpus is to the written expression habits of modern Chinese, the more polysyllabic morphemes are adopted, and then the more different the corpus is from modern Chinese expressions, the more difficult it is to understand.

---

[3] http://corpus.njau.edu.cn/
[4] https://github.com/fxsjy/jieba

c) **The hapax legomena**: It is hapax legomena that occurs only once in the text. The number of hapax legomena is it's quantity in the corpus, while the proportion of hapax legomena is the proportion of the number of hapax legomena in the corpus to word type count. According to Popescu and Altmann (2008), hapax legomena can be regarded as an indicator measuring the degree of language analysis, that is, for two texts of the same size, texts with more hapax legomena are richer in terms of vocabulary.

d) **The richness of text content**: The richness of the text content can be measured by *H value* and *R value*.

*H value*: Point $H$ is the point on the Zipf curve there in the distances to the x and y axes are equal. The special feature of this point is that most of the words preceding it are functional words with only grammatical meaning, while most of the words following it are substantives with real semantic meaning. The value corresponding to this point is the $H$ value (Popescu and Altmann, 2006).

*R value*: The $R$ value calculates richness in terms of word frequency and arc length (Popescu and Altmann, 2011). While calculating $R$ value, all words in the corpus are arranged in decreasing order of frequency of word occurrence, with high-frequency words first and low-frequency words second, to obtain a list of word types, and the arc length is the sum of the euclidean distances of the frequencies of two adjacent words in the list. The $R$-value is calculated by the formula:

$$R = 1 - \frac{L_h}{L}$$

The arc length based on point $H$ and the arc length of the entire text $L_h$ and $L$ are respectively calculated as follows:

$$L = \sum_{r=1}^{V-1} \{[f(r) - f(r+1)]^2 + 1\}^{1/2}$$

$$L_h = \sum_{r=1}^{[h]-1} \{[f(r) - f(r+1)]^2 + 1\}^{1/2} + \{(h - f([h]))^2 + (h - [h])^2\}^{1/2}$$

where $f(r)$ represents the frequency of the $r$-th word in the word frequency list. $V$, h and $[h]$ separately represent the number of word type, $H$ value and the value of h rounded upwards. Liu (2018) states that the $R$ value reflects the coverage rate of substantives, explaining that the higher the $R$ value of the corpus is, the higher the proportion of substantives contained is. Whereas the type token ratio and hapax legomena only examine the lexical usage from the perspective of word frequency, $H$ values and the $R$ value draw a boundary between notional words and empty words, contributing to presenting the proportion of substantives in a text. Since notional words have real meanings, texts with more notional words can be considered to contain more things and concepts and are more informative.

(e) **Indicator $a$**: $H$ value is a quantity related to the text length($N$). Indicator $a$ is introduced to remove the effect of text length. The relationship between text length $N$ and point $H$ can be represented by the total area below the rank order curve (Liu and Pan, 2015; Pan, 2015), i.e. $N = aH^2$, then $a$ can be expressed as

$$a = N/H^2$$

Similar to $H$ value and the $R$ value, Martináková et al. (2008) point out that larger values indicate richer text content.

**Sentence Metrics**

It intuitively shows the distance between subordinate words and dominant words, which is a critical indicator of measuring the complexity of the syntactic structure. Some studies on linguistic complexity related to sentence processing mechanism show that when the brain conducts syntactic analysis, the words in the sentence are constantly stored in working memory, and only when the dominant words appear can they be removed. A more significant dependency distance means that more words will linger in working memory. Since the human brain has a limited working memory capacity, it is challenging to understand sentences when the number of words stored exceeds the memory capacity. Hence, a smaller dependency distance reduces the complexity of the syntactic structure, making comprehension easier (Gildea and Temperley, 2010).

Aiming to compare the complexity of different sentences, Liu (2009) proposes that the dependency distance of a sentence containing multiple dependencies can be denoted as the average of the dependency distances, i.e. the dependency distance of a sentence of length n is:

$$MDD = \frac{1}{n-1}\sum_{i=1}^{n-1}|DD_i|$$

Where $DD_i$ is the ith dependency distance in a sentence. the dependency distance is the average dependency distance of all sentences in the article. we employ the natural language processing tool pyltp[5] released by Harbin Institute of Technology to label the dependency relationship to obtain the dependency relationship of Chinese news text.

## Research Results

**Vocabulary Measurement Results**

Regarding the six indicators mentioned in the section "Lexical Metrics" as evaluation metrics, we have counted the Chinese vocabulary in each of the six countries, and the results are shown in Table 2. Moreover, as the total size of texts from different sources differs, we also conduct a Spearman correlation test between each indicator and text size separately to eliminate the effect of text size. The results reveal that there is no significant correlation between the indicators and text size. Thus, we can make a direct comparison between these indicators of Chinese lexicons from different nations.

We can see from the statistics in table 2 that:

Chinese news vocabulary in the Philippines has the highest type token ratio of 0.237, while Chinese in China has the lowest type token ratio of 0.160. All of the Philippines, Malaysia and Singapore have type token ratio higher than 0.20, which suggests that these countries are richer in the usage of words concerning Chinese news reports.

In terms of hapax legomena, the hapax legomena proportion of Chinese in Philippines is highest, followed by Singapore, Malaysia and China, while Chinese news in Indonesian accounts for lowest proportion in hapax legomena. The result is mostly consistent with previous results of the type token ratio, which further suggests that Chinese news in the Philippines has the highest lexical richness, and that Singapore and Malaysia also tend to select a wider variety of words when reporting the event using Chinese. In contrast, Indonesia and Brunei have a more limited range of vocabulary used in their Chinese news. In Appendix 1, we enumerate 100 hapax legomenas randomly selected from the Chinese language corpus in each nation. The observations demonstrate that the occurrences of the hapax legomena can be classified into the following categories:

(1) Idioms: Chinese language reports in each nation of the Malay Archipelago tend to use idioms, most of which are used in a limited number of scenarios and therefore occur less frequently. What's more, Indonesia and Singapore account for the highest proportion of idioms, both reaching 12%.

(2) Proper nouns: Proper nouns, for instance, names of people and places, do not appear repeatedly in different reports.

(3) Common words: The limited size of our corpus has led to the low frequency of some common words.

---

[5] https://github.com/HIT-SCIR/pyltp

**Table 2 Vocabulary measurement results**

| Country | Philippines | Indonesia | Malay | Brunei | Singapore | China | Relevance |
|---|---|---|---|---|---|---|---|
| Number of word type | 10407 | 7677 | 9977 | 8746 | 11451 | 9830 | 0.09(0.872) |
| Type token ratio | 0.237 | 0.177 | 0.204 | 0.176 | 0.233 | 0.160 | -0.6(0.208) |
| Number of monosyllabic free morphemes | 894 | 714 | 671 | 643 | 1251 | 1236 | 0.143(0.787) |
| Proportion of monosyllabic free morphemes | 8.59% | 9.30% | 6.73% | 7.35% | 10.92% | 12.57% | -0.714(0.111) |
| Number of hapax legomena | 5839 | 3956 | 5359 | 4555 | 6387 | 5207 | 0.086(0.872) |
| Proportion of hapax legomena | 56.11% | 51.53% | 53.7% | 52.08% | 55.78% | 52.97% | -0.029(0.957) |
| $H$ value | 61 | 73 | 74 | 78 | 66 | 69.5 | 0.2(0.704) |
| The $R$-value | 0.982 | 0.975 | 0.980 | 0.975 | 0.983 | 0.979 | -0.029(0.957) |
| $a$ | 11.78 | 8.16 | 8.93 | 8.16 | 11.29 | 10.44 | 0.029(0.957) |

The $R$ value: The difference in the $R$ value between the Chinese texts of the six countries is not significant, all being above 0.975. Among them, the R value of Chinese texts in Singapore is the highest, reaching 0.983, followed by the Philippines and Malaysia. The $R$ value for Brunei and Indonesia is lower than the $R$ value for the People's Daily text, but the difference is quite small. And the result is also largely consistent with the previous results for the hapax legomena. It can be seen that Singapore, Philippines and Malaysia use richer vocabulary and refer to more things or concepts containing more content when reporting news in Chinese.

$a$ value: As can be seen, the final $a$ value further corroborates the results reflected by the previous

R-value. Chinese news in the Philippines possesses the highest textual richness, and both the Philippines and Singapore also tend to convey more information when reporting in Chinese compared to China. In contrast, the textual richness of Chinese news in Indonesia and Brunei is lower.

Monosyllabic free morphemes: Chinese language in Singapore, Indonesia and Philippines account for the highest proportion of monosyllabic free morphemes, with Singapore having over 10% of monosyllabic free morphemes, indicating that the Chinese language of Singapore, Indonesia and Philippines are the most challenging to understand in terms of comprehension difficulty.

In summary, compared to the Mainland, the Philippines, Malaysia and Singapore have higher content richness among Chinese news in the five nations. In terms of text complexity, Chinese news texts in Singapore, Indonesia and the Philippines are most complex and incomprehensible.

**Sentence Measurement Results**

We calculate the average dependency distance of Chinese news texts in each country separately, as shown in Table 3. Smaller dependency distance is conducive to reducing the complexity of syntactic structures, which makes it easier to understand for readers. According to the results, the average dependency distance of Chinese news texts from Indonesia, Malaysia and Brunei is relatively large. It can be considered that the sentences are more complex in structure, which may spend some time for the readers to clarify the sentence structure. In Singapore and the Philippines, the average dependency distance of Chinese news texts is smaller than that of the Mainland, intuitively meaning that readers enable to understand the sentences more quickly and get through the whole article.

We further select three sentences with high dependency distances in the Chinese text of each country, as shown in Table 4. The majority of the sentences with high dependency distances are found to be in prose form, or enumerate multiple specific items, or contain multiple clauses, i.e. one sentence contains multiple pieces of information. These pose a challenge to the memory of the human brain, resulting in more difficulty to understand the sentences.

Table 3 Results of sentence measurement

| Country | Philippines | Indonesia | Malaysia | Brunei | Singapore | China |
| --- | --- | --- | --- | --- | --- | --- |
| Average dependency distance | 3.716 | 4.002 | 3.980 | 4.217 | 3.863 | 3.975 |

Table 4 Examples of sentences with larger dependency distances in each country

| Country | Dependency distance | Sentences |
| --- | --- | --- |
| Indonesia | 7.69 | 参加 Gunungsitoli 码头自动门控制系统启用仪式的有 Gunungsitoli 市市长 Ir.Lak-homizaroZebua, 尼亚斯警分署 BazawatoZebua,SH,MH 助理大警监、Gunungsitoli 市国家法院院长和 Gunungsitoli 码头 KSOP 执行主任 Rochadi,SE. |
| | | 印尼第一码头公司下辖的 Gunungsitoli 码头总经理 AuliaRahmanHasibuan,SE.MM 在致辞中说，Gunungsitoli 码头使用自动门控制系统，是政府在提升码头管理程序环节中的一项突破。
*(The opening ceremony of the automatic gate control system at Gunungsitoli Terminal was attended by the Mayor of Gunungsitoli Municipality, Ir.Lak-homizaroZebua, Assistant Chief Superintendent of Police BazawatoZebua, SH, MH of Nias Police Station, the President of the National Court of Gunungsitoli Municipality and the KSOP Executive Director Rochadi of Gunungsitoli Terminal. In the speech, AuliaRahmanHasibuan, SE.MM, General Manager of Gunungsitoli Terminal under the First Indonesian Terminal Company, said that the usage of the automatic gate control system at Gunungsitoli Terminal is a breakthrough in upgrading the terminal management process.)* |
| | 8.79 | 中央农办主任韩俊说，"文件坚持问题导向，突出统筹推进农村经济建设、政治建设、文化建设、社会建设、生态文明建设和党的建设，加快推进乡村治理体系和治理能力现代化，加快推进农业农村现代化，走中国特色社会主义乡村振兴道路，是谋划新时代乡村振兴的顶层设计。"
*(Han Jun, director of the Central Agricultural Office, said, "The document insists on being problem-oriented, highlights the integrated promotion of rural economic construction, political construction, cultural construction, social construction, ecological civilization construction and Party construction, accelerates the modernization of the rural governance system and governance capacity, accelerates the modernization of agriculture and rural areas, and follows the path of socialist rural revitalization with Chinese characteristics, and is a top-level design for planning rural revitalization in the new era.")* |
| | 8.80 | 周三，Bapepam-LK 会计和信息公开准则局长 EttyRetnoWulandari 与印尼会计师协会在雅京苏迪曼中央商业区 RitzCalton 酒店的 PasificPlace 厅召开有关推广上述 13 条财务会计准则会议中，如此披露。
*(On Wednesday, so disclosed EttyRetnoWulandari, Director General of Accounting and Disclosure Standards of Bapepam-LK, during a meeting promoting the above 13 financial accounting standards with the Indonesian Institute of Accountants at the PasificPlace Hall of the Ritz-Carlton Hotel in Yakin Sudirman Central Business District on Wednesday.)* |
| Malay | 7.37 | 李克强总理在赶赴灾区飞机上召开会议，听取汇报后作出 7 项部署：提紧时机救人，严密防范灾害发生，加强救治，做好灾民安置，统一指挥，做好恢复生产准备，最后是着手恢复重建工作；要快速而有序，救灾及救援要同步推进， |
| | | 并要根据灾区需求，建立以地方为主的统一指挥体系，中央各部门做好协调和保障。
*(Premier Li Keqiang held a meeting on the plane rushing to the disaster area, listened to the report and made seven deployments: to seize the opportunity to save people, to closely prevent the occurrence of disasters, to strengthen the rescue and treatment, to perform well in resettling the victims, to unify the command, to prepare for the resumption of production, and finally to start the recovery and reconstruction work; to be fast and orderly, disaster relief and rescue should be promoted simultaneously, and to establish a unified command system based on the needs of the disaster area, with the central departments doing a good job of coordination and protection.)* |
| | 7.54 | 另一方面，针对其将在何时会见首相拿督斯里阿都拉和国阵野新区国会议员拿督莫哈末赛益的问题时，也是国阵党鞭的纳吉表示，有关的日期还没有决定。
*(On the other hand, in response to a question on when he would meet Prime Minister Datuk Seri Abdullah and BN MP for the new wilderness district Datuk Mohamed Sayik, Najib, who is also the BN party whip, stated that the dates had not yet been decided.)* |
| | 7.56 | 班迪卡阿敏强调，他无需向在首相兼财长拿督斯里纳吉在10月21日提呈财政预算案时离席抗议的在野议员道歉。
*(Bandekar Amin stressed that he did not need to apologize to the opposition MPs who left the meeting in protest when Prime Minister and Finance Minister Datuk Seri Najib presented the Budget on October 21.)* |
| Philippines | 6.11 | 她说菲律宾红十字会有近200万想成为器官捐献者的志愿者，"我们要鼓励我们的员工，使无偿献血成为生活的一种方式，同时当它涉及到生命救助时，还有其他的需求，而不仅仅只有血液，有人需要器官"。
*(She said that the Philippine Red Cross has nearly two million volunteers who would like to become organ donors and "we want to encourage our staff to make blood donation a way of life, and when it comes to life-saving, there are other needs, not just blood, but organs".)* |
| | 8.94 | 由于异地借考需要对考生户籍相关情况进行严格审核,而对这类户籍原在外省、刚刚迁入不久、又没有迁到广州的考生，广州本地的招考部门很难对其户籍的详细、真实情况进行核实，故出此政策。
*(This policy has been introduced because it is difficult for the local recruitment and examination authorities in Guangzhou to verify the details and authenticity of a candidate's household registration, which was originally in a foreign province and has just moved to Guangzhou.)* |

| | | | |
|---|---|---|---|
| | 8.98 | 央行说，"非居民在债券(或海外母公司放贷给它们的本地子公司以融资现有运作和商业扩张)的投资较一年前的二亿五千四百万美元增加百分之一百二十二点六，达到五亿六千六百万美元。" | |
| | | *(The central bank said that "investments by non-residents in bonds (or loans by overseas parent companies to their local subsidiaries to finance existing operations and business expansion) rose by 122.6 per cent to $566 million from $254 million a year ago.")* | |
| Singapore | 7.94 | 第一期强棒出击，邀请韩国第一代男子团体 g.o.d 担任嘉宾，虽然少了尹启相，但当其他四名成员以当年形象登场时，还是掀起现场一片欢腾。已是大叔级的 g.o.d，跳着当年的舞蹈，逗嗨全场。 | |
| | | *(In the first episode, the first Korean male group g.o.d was invited as a guest. Although Yoon Ki-sang was absent, the other four members of the group appeared in their old image and created a lot of excitement. The four members of g.o.d, who are now older men, danced in the same way they did back in the day and entertained the audience.)* | |
| | 8.04 | 目前，祖基菲利与马来青年乐团正致力探索甘美兰音乐，创作一系列新作品。尽管年纪轻轻，黄淑敏（30 岁）在剧场界早已凭"wo(men)"（2010）、"ForBetterorForWorse"（2013）、"Normal"（2015）等原创剧作声名鹊起，也曾与必要剧场、戏剧盒和滨海艺术中心合作。 | |
| | | *(Currently, Zukifili and the Malay Youth Orchestra are exploring Gamelan music and creating a series of new works. Despite her young age, Wong Suk Min (30) has already made a name for herself in the theatre world with original productions such as "wo(men)" (2010), "ForBetterorForWorse" (2013) and "Normal" (2015), and has also worked with Essential Theatre, Drama Box and Esplanade.)* | |
| | 8.63 | "康熙"花 10 分钟找默契，小 S 新秀不豪放不揩油，台湾艺人蔡康永和小 S（徐熙娣）相隔两年，再度联手主持中国大陆网络节目《真相吧！花花万物》，首集来宾邀请大陆综艺天后谢娜，三天流量破亿，可见"康熙"魅力不减。 | |
| | | *(The new show of "Kangxi" took 10 minutes to find a tacit understanding and Xiao S is not a boozy or molesting show. Taiwanese artistes Tsai Kang-yong and Xiao S (Xu Xiti) joined forces again after two years to host the Mainland network show 《Truth!》. The first episode of the show featured mainland variety diva Xie Na, and the traffic flow broke 100 million in three days, showing that the charm of "Kangxi" is still intact.)* | |
| Brunei | 7.36 | 该校董事长蔡仔运在致辞中提到说学校将会进行大装修，重建新的教学建筑，明年学生将会迁移到临时教室上课。临时教室已经通过救拯局以及卫生局的批 | |

| | | 准，确保学生们可以在安全和卫生的环境里上课。同时他也非常感谢新顺利所赞助的临时教室，让孩子们可以在原来的校址上课，不用迁到其他地方。最后，董事长也吁请大家继续大力支持中岭学校。 |
|---|---|---|
| | | *(In his speech, the school's chairman, Mr. Chua Chai Wan, said that the school would be undergoing a major renovation to rebuild the new building and that students would be relocated to temporary classrooms next year. The temporary classrooms have already been approved by the Rescue Department and the Health Department to ensure that the students can attend classes in a safe and hygienic environment. He also thanked New Smooth for sponsoring the temporary classrooms so that the children attend classes at the original school site instead of moving elsewhere. In closing, the Chairman urged everyone to continue supporting Chung Ling School.)* |
| | 7.69 | (汶莱斯市 27 日讯)汶莱社会发展局为减轻孤儿们在来临开斋节的经济负担，今早在汶莱国际会议中心举行"2015 年颁发孤儿佳节辅助金大会"，今年共有 5213 名孤儿领取辅助金。 |
| | | *((Brunei 27) - The Brunei Social Development Board (BSDB) held the "2015 Orphan's Festive Aid Presentation" at the Brunei International Convention Centre this morning to ease the financial burden of orphans in the run-up to Eid.)* |
| | 7.81 | 此外，丕显拿督哈芝阿都阿兹斯也在教育事业合作作出努力，特别是在 1984 年，1986 年，1988 年至 1992 年，1994 至 1998 年和 2000 至 2004 年期间，在东南亚教育部长组织区域委员会里贡献良多。 |
| | | *(In addition, Phi Hsien Dato Hazi Abdul Aziz has made a tremendous contribution to cooperation in education, particularly in the SEAMEO Regional Committee in 1984, 1986, 1988-1992, 1994-1998 and 2000-2004.)* |

**Table 5 The five dependencies with the highest percentage of Chinese language in each country**

| | Philippines | | Indonesia | | Malaysia | | Brunei | | Singapore | | China | |
|---|---|---|---|---|---|---|---|---|---|---|---|---|
| Type | Dis. | Prop. | Dis. | Prop. | Dis. | Prop. | Dis. | Prop. | Dis. | Prop. | Dis. | Prop. |
| ATT | 1.83 | 0.25 | 1.85 | 0.31 | 1.78 | 0.26 | 1.97 | 0.36 | 1.87 | 0.22 | 1.86 | 0.27 |
| WP | -6.83 | 0.17 | -6.05 | 0.16 | -6.44 | 0.16 | -7.75 | 0.13 | -6.45 | 0.19 | -7.53 | 0.16 |
| ADV | 3.49 | 0.14 | 4.15 | 0.12 | 3.41 | 0.14 | 4.30 | 0.10 | 2.91 | 0.14 | 3.43 | 0.12 |
| VOB | -3.58 | 0.10 | -4.02 | 0.10 | -3.69 | 0.12 | -4.47 | 0.09 | -3.22 | 0.11 | -3.72 | 0.10 |
| COO | -8.54 | 0.09 | -7.79 | 0.09 | -7.60 | 0.09 | -8.68 | 0.09 | -7.95 | 0.10 | -8.38 | 0.10 |

Moreover, we show in Table 5 the five most frequent dependencies in the texts of each country. Interestingly, the top five dependency relationships are all ATT (Attribute), WP (Punctuation), ADV (Adverbial), VOB (Verb-object) and COO (Coordinate). As can be seen, the Chinese language in the countries of the Malay Archipelago will make extensive usage of the "attributive + head" in sentences, even exceeding the usage of punctuation (i.e. WP). Among them, Indonesia and Brunei, use the "attributive + head" structure more frequently, with ATT accounting for more than 0.3. These differences can reflect the fact that in terms of the organization of sentence structure, Mainland Chinese and Malaysian Chinese have developed more distinct differences as they have developed in their respective territories.

**Feature Words**

Liu indicates that there are words with Southeast Asian characteristics in Southeast Asian media texts (Liu, 2010; 2011). In addition to the quantitative research on the text, we also attempt to extract words with large differences between Chinese of each country and Chinese of the Mainland in the corpus, namely feature words. For this purpose, we adopt pre-filtering, supplemented by manual handling. Initially, we construct the list of word types regarding the corpus from *People's Daily*. Then for the Chinese text of each nation, the words appearing in the list are removed. Subsequently, the pre-filtered texts are further filtered manually. Finally, 56 Chinese feature words for the Philippines, 21 Chinese feature words for Indonesia, 21 Chinese feature words for Malaysia, 23 Chinese feature words for Brunei and 14 Chinese feature words for Singapore were collated (see Appendix 2 for details). After extracting and classifying the feature words, it is found that they can be broadly divided into three categories:

(a) It is the same as the Mainland terminology, except that individual words are written differently.
(b)There are some differences, but it is still relatively easy to see the meaning of the words;
(c)There are significant differences, and it is difficult to understand only by reading the words.

The majority of the Chinese in the nations of the Malay Archipelago are immigrants and the descendants migrating from Fujian, Guangdong, Guangxi, Hainan and Taiwan since the Ming and Qing dynasties. Owing to the lack of communication with the Mainland, all diverge from their modern Chinese mainland counterpart with centuries of inheritance and development. Numerous words still retain ancient or variant characters, for instance, "菲律宾" is written in Chinese as "菲律滨" in the Philippines, and "咨询" is written in Chinese as "谘询" in Malaysian. The seepage force of the dialects cannot be ignored. In an environment where various dialects such as Minnan, Cantonese and Hakka are prevalent, the Chinese language of the Malay Archipelago has absorbed many dialect features, such as the isomorphic reverse words( "净洁"与"洁净" ) that frequently occur in these dialects. Many dialect words originally used only in the spoken language have been widely accepted and employed in the written language, e.g. "冻未条" in the Minnan dialect. In addition, massive words with regional characteristics occur influenced by social cultures, such as the unique word "报生纸" in Malaysian Chinese, referring to the certificate issued by the government to certify the birth of a Malaysian.

## Conclusion

In this paper, we conducted a quantitative analysis of Chinese news texts collected from five Malay Archipelago nations, trying to figure out their differences from the texts written in modern standard Chinese, from a lexical and syntactic perspective. The results reveal that the Chinese texts of the Philippines, Malaysia and Singapore are richer in vocabulary and content than those in Indonesia and Brunei from the lexical perspective, while the Chinese texts of Indonesia and Brunei have a more limited vocabulary scope and relatively less information. In terms of sentences, the mean dependency distance of Chinese news texts collected from Indonesia, Malaysia and Brunei is higher, meaning more complex sentence structure; in contrast, Chinese news texts from Singapore and the Philippines are easier to understand and read. The measurement indicators suggest that the development of Chinese in the nations of the Malay Archipelago is different, while all diverge from the modern Chinese mainland. Meanwhile, we tentatively extract and classify several featured Chinese words in five nations and briefly analyze the causes. We attempt to further explore whether the Chinese language in the Malay Archipelago shows such differences from that in the mainland in other genres as well.

## Declarations

**Funding**

This work was supported by the Key Field Project for Universities of Guangdong Province (No. 2019KZDZX1016), the National Natural Science Foundation of China (No. 61572145) and the National Social Science Foundation of China (No. 17CTQ045).

**Conflicts of interests**

The authors declare that we do not have any commercial or associative interest that represents a conflict of interest in connection with the work submitted and that the research do not involve human participants and/or animals.

**Appendix 1** Examples of hapax legomena in Chinese in various countries

Table 6 100 hapax legomenas of Indonesian Chinese

| 陈美致 | 异性 | 好话 | 施工 | 廖省 |
|---|---|---|---|---|
| 林惠宽 | 春种 | 维吉尼亚州 | 巴东 | 北美 |
| 张少杰 | 毫厘之差 | 冰风暴 | 浑身解数 | 哈桑 |
| 布面 | 显舍 | 连通性 | 孟欣 | 不择手段 |
| 航点 | 复苏期 | 联合体 | 农办 | 摩纳斯 |
| 伍拉莱 | 搁 | 黄世章 | 危机四伏 | 指示器 |
| 凶煞 | 碳氧化物 | 丹格朗县 | 雨后春笋 | 特型 |
| 司空见惯 | 专门 | 虎视眈眈 | 兜售 | 嫌犯 |
| 冒毽 | 握手 | 职业 | 纺织业 | 湖人队 |
| 开元寺 | 区别 | 林汉忠 | 异口同声 | 艰巨 |
| 气血 | 折算 | 搁风 | 凶曜 | 肯塔基州 |
| 留影 | 同胞 | 马长山 | 嫌疑 | 专科 |
| 耕作业 | 贪图 | 附加费 | 俾 | 回函 |
| 总领事 | 复兴 | 旅游部 | 古邦县 | 揭晓 |
| 里约 | 如此一来 | 衡量 | 时至今日 | 化妆品 |
| 弃权 | 环节 | 递升 | 不得而知 | 良机 |
| 眼前 | 落选 | 4月份 | 栏目 | 一个个 |
| 导航 | 亟待解决 | 作弊 | 家谱 | 甲骨文 |
| 圣地 | 全心全意 | 远渡 | 夫君 | 毒贩 |
| 个别 | 撰写 | 坚决 | 匮乏 | 刁难 |

Table 7 100 hapax legomenas of Malaysian Chinese

| 举世 | 林耿光 | 建筑师 | 察看 | 西连达奴 |
|---|---|---|---|---|
| 受困 | 反应 | 浮刹 | 停靠 | 圣女 |
| 合力 | 裁员 | 煤气管 | 地基 | 都门 |
| 古邦 | 警署 | 圣日曼 | 黄圣南 | 辅仁 |
| 校董 | 标普 | 废墟 | 哥姆士 | 案率 |
| 卧尸 | 谢依思 | 海口区 | 张清 | 承受 |

| 斋戒 | 一阵子 | 钜细靡遗 | 德维道维兹 | 喷发 |
| --- | --- | --- | --- | --- |
| 长途 | 全村 | 族群 | 不以为意 | 科学局 |
| 徒劳无功 | 曼吉斯 | 雅加达 | 王耀新 | 罪案 |
| 医药费 | 质疑 | 房静琳 | 残障者 | 分道扬镳 |
| 蠕虫 | 竖立 | 外籍 | 胡嘉华 | 叶秀美 |
| 梅威瑟 | 造访 | 锦标赛 | 天然气 | 弹簧刀 |
| 情人节 | 简讯 | 往来 | 锁带 | 苏牙 |
| 追溯 | 赛哈密 | 洪春本 | 郎朗 | 难以置信 |
| 张泰卿 | 泰晤士报 | 胜率 | 卫星 | 戏剧社 |
| 入狱 | 眼神 | 部落客 | 关税局 | 杨哲宁 |
| 网媒 | 勒索 | 罗正棠 | 据称 | 孜孜不倦 |
| 胃病 | 牛津 | 弗内斯 | 达雅族 | 响彻 |
| 按钮 | 陈祺荃 | 墨西哥城 | 在野 | 争论性 |
| 外围 | 老人院 | 重金 | 火种 | 劝告 |

Table 8 100 hapax legomenas pf Filipino Chinese

| 早先 | 协调员 | 波帕纳紧 | 华尔街 | 鼎立 |
| --- | --- | --- | --- | --- |
| 其馀 | 菲国 | 开除令 | 焦虑 | 陶瑞明 |
| 用俱 | 荷兰 | 缉毒署 | 投标者 | 趋炎附势 |
| 星期二 | 艰巨 | 承购者 | 刚正不阿 | 高考生 |
| 崇高 | 罗素静 | 加勒比 | 宿务省 | 立陶宛 |
| 南美 | 低谷 | 狙击手 | 邦义利兰 | 微观 |
| 防御 | 北怡洛戈 | 召集 | 评分 | 冰岛 |
| 政党 | 市值 | 雪上加霜 | 邢文华 | 大都会 |
| 袁泉 | 许昌 | 大咖 | 革新 | 备料 |
| 繁荣 | 司示部 | 寡照 | 特许权 | 加牙渊省 |
| 会考 | 破除 | 6亿 | 摊位 | 辩论 |
| 两行 | 私语 | 摄氏度 | 微信 | 补救 |
| 七情六欲 | 出品 | 整理 | 夜幕 | 国际法 |
| 无量 | 无可奈何 | 不得已 | 纸钱 | 质感 |

| 挤占 | 自豪 | 金俞星 | 终极 | 祈福 |
| 方便 | 转角 | 阶层 | 收手 | 红点 |
| 董先生 | 劝住 | 管豁 | 盼 | 公私合营 |
| 南方日报 | 惠顾 | 惘然 | 台账 | 愤怒 |
| 体校 | 进取心 | 踩杀 | 颠覆性 | 中粮 |
| 关联 | 庇护 | 局势 | 交头接耳 | 影片 |

Table 9 100 hapax legomenas of Singapore Chinese

| 坊间 | 王嘉尔 | 台大 | 阿卜杜勒 | 壁橱 |
| 锻炼 | 卖命 | 张书豪 | 所在地 | 顶限 |
| 浏海 | 笑口常开 | 兰开斯特 | 剪辑师 | 古曲 |
| 艾米汉默 | 蒂娜 | 鲁妮玛拉 | 排斥 | 懵懂 |
| 乐坛 | 出炉 | 埃迪雷德梅尼 | 对讲机 | 征召 |
| 原著 | 潮流 | 设身处地 | 赞誉 | 咖啡 |
| 高雅罗 | 修复 | 剧团 | 李帝勋 | 蘑菇 |
| 荷兰 | 星期五 | 吉打州 | 马尼拉 | 配搭 |
| 老戏骨 | 心知肚明 | 首席 | 计算 | 不了了之 |
| 淡马锡 | 赞不绝口 | 肾上腺素 | 参咨 | 高云翔 |
| 混合体 | 京都 | 皮夹 | 显露 | 渡过 |
| 荣登 | 莫小玲 | 惧高症 | 重头戏 | 留宿 |
| 榜上有名 | 考取 | 涉猎 | 首付 | 温文尔雅 |
| 小孩子 | 获准 | 国医 | 腹肌 | 惠英红 |
| 支支吾吾 | 巧思 | 珠宝行 | 截足灭证 | 发薪 |
| 眼明手快 | 精湛 | 谦卑 | 事宜 | 鱼皮 |
| 诗作 | 惊疑 | 卢广仲 | 前所未有 | 舞剧 |
| 涌出 | 千方百计 | 严审 | 克服 | 领结 |
| 释怀 | 演过 | 粤语 | 发明 | 理事会 |
| 反常 | 力求 | 手气 | 恐惧 | 激励 |

Table 10 100 hapax legomenas of Brunei Chinese

| 致力 | 推介礼 | 单身 | 卡威恩阿斯里 | 马来熊 |

| 就职 | 优胜奖 | 吴美燕 | 乙氨基酚片 | 艾文 |
| 毗邻 | 奖励 | 耶雅 | 希拉克 | 伊扎特 |
| 待业 | 文化处 | 侨胞 | 月份 | 优质 |
| 榕籍 | 宣导会 | 审计局 | 呕吐 | 营销 |
| 哈芝巴林 | 活跃期 | 阿尼斯拉 | 双溪都 | 标示 |
| 悠闲 | 大卫坎贝尔 | 赝制 | 心脏病 | 哥斯达黎 |
| 病例 | 马华迪 | 台北 | 内安局 | 艺术处 |
| 送行 | 黄义明 | 恶意 | 濒临 | 沟通 |
| 事宜 | 地主 | 受宠若惊 | 忍无可忍 | 坐享其成 |
| 林国钦 | 夺走 | 佰都利卡 | 对像 | 多元文化 |
| 本基兰阿纳曼 | 熟练 | 救济 | 教育局 | 商务部 |
| 收取 | 2006 年 | 寿司 | 陈丽珍 | 奖予 |
| 俄国 | 审查 | 灾民 | 签字 | 25 份 |
| 预览 | 采摘 | 办房 | 里占林葛 | 哀伤 |
| 保释 | 超速 | 冲击 | 被弃 | 不遗余力 |
| 1 间 | 追踪 | 差错 | 帐号 | 密集 |
| 留学生 | 判决 | 礼仪 | 罗斯阿米娜 | 史无前例 |
| 接力赛 | 林国民 | 反锁 | 包容性 | 书画 |
| 1933 年 | 查账 | 胶带 | 校方 | 缆线 |

**Appendix 2** Feature words of Chinese language in various countries

Table 11 Feature words of Filipino Chinese language

| Category | Philippine | Mainland | Philippine | Mainland | Philippine | Mainland |
| --- | --- | --- | --- | --- | --- | --- |
| a | 迅息 *(message)* | 讯息 *(message)* | 覆审 *(review)* | 复审 *(review)* | 担纲 *(serving as)* | 担任 *(serving as)* |
| | 扫瞄 *(scan)* | 扫描 *(scan)* | 婚喜 *(wedding)* | 婚宴 *(wedding)* | 声请 *(application)* | 申请 *(application)* |
| | 保镳 *(bodyguard)* | 保镖 *(bodyguard)* | 坦承 *(honesty)* | 坦诚 *(honesty)* | 埋单 *(pay the bill)* | 买单 *(pay the bill)* |
| | 英人 *(British)* | 英国人 *(British)* | 义大利 *(Italy)* | 意大利 *(Italy)* | 超渡 *(transcendence)* | 超度 *(transcendence)* |

|   | 菲律滨<br>*(Philippines)* | 菲律宾<br>*(Philippines)* | 形像<br>*(profile)* | 形象<br>*(profile)* | 臭豆付<br>*(stinky tofu)* | 臭豆腐<br>*(stinky tofu)* |
|---|---|---|---|---|---|---|
|   | 想像<br>*(imagine)* | 想象<br>*(imagine)* | 渡假村<br>*(resort)* | 度假村<br>*(resort)* | 转瞬即届<br>*(pass in a flash)* | 转瞬即逝<br>*(pass in a flash)* |
|   | 座落<br>*(situated)* | 坐落<br>*(situated)* | 亚州<br>*(Asia)* | 亚洲<br>*(Asia)* | 手板电脑<br>*(tablet computer)* | 平板电脑<br>*(tablet computer)* |
|   | 付帐<br>*(pay the bill)* | 付账<br>*(pay the bill)* | 游漓<br>*(free from)* | 游离<br>*(free from)* | 走头无路<br>*(no way out)* | 走投无路<br>*(no way out)* |
|   | 谘询<br>*(inquiry)* | 咨询<br>*(inquiry)* | 义大利<br>*(Italy)* | 意大利<br>*(Italy)* | 用俱<br>*(use of tools)* | 用具<br>*(use of tools)* |
|   | 躬临<br>*(visit)* | 莅临<br>*(visit)* | 席次<br>*(seat)* | 席位<br>*(seat)* | 情资<br>*(intelligence)* | 情报<br>*(intelligence)* |
|   | 协易<br>*(agreement)* | 协议<br>*(agreement)* | 示称<br>*(claim)* | 声称<br>*(claim)* | 救拔<br>*(redemption)* | 救赎<br>*(redemption)* |
|   | 棺罩<br>*(coffin)* | 棺材<br>*(coffin)* |   |   |   |   |
| b | 馀地<br>*(leeway)* | 余地<br>*(leeway)* | 馀孽<br>*(remnant)* | 余孽<br>*(remnant)* | 职志<br>*(aspiration)* | 志向<br>*(aspiration)* |
|   | 管豁<br>*(jurisdiction)* | 管辖<br>*(jurisdiction)* | 罚锾<br>*(fines)* | 罚款<br>*(fines)* | 提捐<br>*(donation)* | 捐赠<br>*(donation)* |
|   | 祇要<br>*(as long as)* | 只要<br>*(as long as)* | 其馀<br>*(the rest)* | 其余<br>*(the rest)* | 爱载<br>*(love and adore)* | 爱戴<br>*(love and adore)* |
|   | 呎<br>*(foot)* | 尺<br>*(foot)* | 奖获<br>*(award)* | 获奖<br>*(award)* | 马来西亚<br>*(Malaysia)* | 马来西亚<br>*(Malaysia)* |
|   | 变交<br>*(exchange)* | 交换<br>*(exchange)* | 净洁<br>*(clean)* | 洁净<br>*(clean)* | 乘搭<br>*(travel by)* | 搭乘<br>*(travel by)* |
| c | 纾懈<br>*(relieve)* | 疏解<br>*(relieve)* | 科处<br>*(penalty)* | 处罚<br>*(penalty)* | 一途生意<br>*(a single business)* | 一单生意<br>*(a single business)* |
|   | 头寸<br>*(payment)* | 款项<br>*(payment)* | 棺樟<br>*(coffin)* | 棺材<br>*(coffin)* | 冻未条<br>*(can't stand)* | 受不了<br>*(can't stand)* |
|   | 以…而业<br>*(in terms of)* | 以…而言<br>*(in terms of)* |   |   |   |   |

**Table 12** Feature words of Indonesian Chinese language

| Category | Indonesian | Mainland | Indonesian | Mainland | Indonesian | Mainland |
|---|---|---|---|---|---|---|
| a | 课以<br>*(lesson to)* | 科以<br>*(lesson to)* | 致达<br>*(expression)* | 表达<br>*(expression)* | 达致<br>*(achievement)* | 达成<br>*(achievement)* |
|   | 助佑<br>*(bless)* | 保佑<br>*(bless)* | 助动<br>*(assistance)* | 助力<br>*(assistance)* | 重蹈复辙<br>*(repeat the same mistake)* | 重蹈覆辙<br>*(repeat the same mistake)* |
|   | 办结<br>*(process)* | 办理<br>*(process)* | 官非<br>*(lawsuit)* | 官司<br>*(lawsuit)* | 蜜锣紧鼓<br>*(lit working closely on a gong and drumming)* | 密锣紧鼓<br>*(lit working closely on a gong and drumming)* |
|   | 大部<br>*(most)* | 大部分<br>*(most)* | | | | |
| b | 每<br>*(every)* | 每<br>*(every)* | 达<br>*(reach)* | 达<br>*(reach)* | 搧<br>*(fan)* | 扇<br>*(fan)* |
|   | 获奖<br>*(award)* | 获奖<br>*(award)* | 时间<br>*(time)* | 时间<br>*(time)* | 时<br>*(inch)* | 寸<br>*(inch)* |
|   | 启开<br>*(open)* | 开启<br>*(open)* | | | | |
| c | 某图<br>*(a picture)* | 牟图<br>*(a picture)* | 致达<br>*(expression)* | 表达<br>*(expression)* | 特出<br>*(special)* | 特别<br>*(special)* |
|   | 件事<br>*(process)* | 处理<br>*(process)* | | | | |

**Table 13** Feature words of Malaysian Chinese language

| Category | Malaysia | Mainland | Malaysia | Mainland | Malaysia | Mainland |
|---|---|---|---|---|---|---|
| a | 案展<br>*(case)* | 案件<br>*(case)* | 勤力<br>*(diligent)* | 勤劳<br>*(diligent)* | 谘询<br>*(consultation)* | 咨询<br>*(consultation)* |
|   | 款捐<br>*(donation)* | 捐款<br>*(donation)* | 德士<br>*(cab)* | 的士<br>*(cab)* | 达致<br>*(achievement)* | 达成<br>*(achievement)* |
|   | 景像<br>*(scenery)* | 景象<br>*(scenery)* | 路经者<br>*(passers-by)* | 途经者<br>*(passers-by)* | 莫需有<br>*(no need to have)* | 莫须有<br>*(no need to have)* |
|   | 当位<br>*(unit)* | 单位<br>*(unit)* | 间中<br>*(middle)* | 中间<br>*(middle)* | 座落<br>*(located at)* | 坐落<br>*(located at)* |
|   | 上载<br>*(upload)* | 上传<br>*(upload)* | | | | |
| b | 醒觉 | 觉醒 | 通膨 | 通胀 | 胥视 | 考虑 |

|  | (awaken) | (awaken) | (inflation) | (inflation) | (consider) | (consider) |
|---|---|---|---|---|---|---|
|  | 入禀<br>(appeal) | 上诉<br>(appeal) |  |  |  |  |
| c | 摆甫士<br>(pose) | 摆pose<br>(pose) | 如末<br>(as at the end) | 如末<br>(as at the end) | 报生纸<br>(birth certificates) | 所有大马人在出生时,政府所发出的证明文件<br>(birth certificates) |
|  | 推事<br>(judge) | 审判官<br>(judge) |  |  |  |  |

Table 14 Feature words of Brunei Chinese language

| Category | Brunei | Mainland | Brunei | Mainland | Brunei | Mainland |
|---|---|---|---|---|---|---|
| a | 进住<br>(enter and garrison) | 进驻<br>(enter and garrison) | 欢渡<br>(celebrate) | 欢度<br>(celebrate) | 视屏<br>(video) | 视频<br>(video) |
|  | 保持捷<br>(porsche) | 保时捷<br>(porsche) | 列常<br>(normal) | 例常<br>(normal) | 形像<br>(image) | 形象<br>(image) |
|  | 干案<br>(commit a crime) | 作案<br>(commit a crime) | 充份<br>(sufficient) | 充分<br>(sufficient) | 晋入<br>(enter) | 进入<br>(enter) |
|  | 现像<br>(phenomenon) | 现象<br>(phenomenon) | 推展<br>(carry out) | 开展<br>(carry out) | 颁赐<br>(award) | 颁发<br>(award) |
|  | 跟据<br>(according to) | 根据<br>(according to) | 充许<br>(allow) | 允许<br>(allow) | 制裁<br>(sanction) | 制裁<br>(sanction) |
|  | 过份<br>(excessively) | 过分<br>(excessively) | 醒觉<br>(vigilance) | 警觉<br>(vigilance) |  |  |
| b | 讵料<br>(unexpectedly) | 岂料<br>(unexpectedly) | 撙节<br>(adjustment) | 调节<br>(adjustment) | 轮贻<br>(tire) | 轮胎<br>(tire) |
|  | 迎迓<br>(greet) | 迎接<br>(greet) |  |  |  |  |
| c | 办房<br>(service staff) | 服务人员<br>(service staff) | 必较<br>(compare) | 比较<br>(compare) |  |  |

Table 15 Feature words of Singapore Chinese language

| Category | Singapore | Mainland | Singapore | Mainland | Singapore | Mainland |
|---|---|---|---|---|---|---|
| a | 械劫案<br>(robbery) | 抢劫案<br>(robbery) | 採用<br>(adopt) | 采用 <br>(adopt) | 峇厘岛<br>(Bali) | 巴厘岛<br>(Bali) |
|   | 坦承<br>(honesty) | 坦诚<br>(honesty) | 模彷<br>(imitation) | 模仿<br>(imitation) | 松解<br>(relaxation) | 松懈<br>(relaxation) |
|   | 德士<br>(cab) | 的士<br>(cab) | 上载<br>(upload) | 上传<br>(upload) | 造形<br>(styling) | 造型<br>(styling) |
|   | 艰钜<br>(daunting) | 艰巨<br>(daunting) | 行劫<br>(robbery) | 打劫<br>(robbery) |   |   |
| b | 配搭<br>(match) | 搭配<br>(match) | 鼙樂團<br>(drum corps) | 鼓乐团<br>(drum corps) |   |   |
| c | 走透透<br>(travel around) | 走遍<br>(travel around) |   |   |   |   |